\documentclass[letterpaper, 10 pt, conference]{ieeeconf}  

\IEEEoverridecommandlockouts                              
                                                          
\usepackage[table, dvipsnames]{xcolor}
\usepackage[pdftex]{graphicx} 
\usepackage{graphicx}
\usepackage[lofdepth,lotdepth,caption=false]{subfig} 
\graphicspath{{figures/}} 
\DeclareGraphicsExtensions{.pdf,.jpg,.png} 
\usepackage{amsmath} 
\usepackage{amsfonts} 
\usepackage{mathtools} 
\usepackage{tikz} 
\usetikzlibrary{shapes,arrows,fit,calc,positioning,automata,backgrounds}
\usepackage{color} 
\usepackage{multirow} 
\usepackage{hhline} 
\usepackage[bookmarks=false]{hyperref} 
\usepackage{enumerate} 
\usepackage{epstopdf} 
\usepackage{dsfont}
\usepackage{siunitx} 
\usepackage{cite}   
\usepackage{diagbox}    
\usepackage{comment}
\usepackage{todonotes}
\usepackage{float}
\usepackage{booktabs}

\usepackage{algorithm}
\usepackage{algpseudocode}

\colorlet{Green1}{green!90!}
\colorlet{Green2}{green!60!}
\colorlet{Green3}{green!40!}
\colorlet{Green4}{green!20!}
\colorlet{Green5}{green!10!}

\definecolor{Bookcolor}{HTML}{00F9DE}

\makeatletter
\def\@citex[#1]#2{\leavevmode
\let\@citea\@empty
\@cite{\@for\@citeb:=#2\do
{\@citea\def\@citea{,\penalty\@m\ }%
\edef\@citeb{\expandafter\@firstofone\@citeb\@empty}%
\if@filesw\immediate\write\@auxout{\string\citation{\@citeb}}\fi
\@ifundefined{b@\@citeb}{\hbox{\reset@font\bfseries ?}%
\G@refundefinedtrue
\@latex@warning
{Citation `\@citeb' on page \thepage \space undefined}}%
{\@cite@ofmt{\csname b@\@citeb\endcsname}}}}{#1}}
\makeatother

\begin{document}

%

\title{Event-based Navigation for Autonomous Drone Racing\\ with Sparse Gated Recurrent Network}

\author{Kristoffer Fogh Andersen$^{*}$, Huy Xuan Pham$^{*}$, Halil Ibrahim Ugurlu and Erdal Kayacan
\thanks{K.F. Andersen, H. X. Pham, H. I. Ugurlu, and E. Kayacan are with Artificial Intelligence in Robotics Laboratory (AiR Lab), Department of Electrical and Computer Engineering, Aarhus University,
        8000 Aarhus C, Denmark
        {\tt\small 201510430@post.au.dk, \{huy, halil, erdal\} at ece.au.dk}}%
\thanks{(*) These authors contributed equally to this work.}
}



%


\maketitle


\begin{abstract}
Event-based vision has already revolutionized the perception task for robots by promising faster response, lower energy consumption, and lower bandwidth without introducing motion blur. In this work, a novel deep learning method based on gated recurrent units utilizing sparse convolutions for detecting gates in a race track is proposed using event-based vision for the autonomous drone racing problem. We demonstrate the efficiency and efficacy of the perception pipeline on a real robot platform that can safely navigate a typical autonomous drone racing track in real-time. Throughout the experiments, we show that the event-based vision with the proposed gated recurrent unit and pretrained models on simulated event data significantly improve the gate detection precision. Furthermore, an event-based drone racing dataset\footnote{The code and data will be available at \url{https://github.com/open-airlab/neuromorphic_au_drone_racing.git}} consisting of both simulated and real data sequences is publicly released.

\end{abstract}
\IEEEpeerreviewmaketitle


\section{Introduction}

The use of event cameras represents a paradigm shift in the way we capture and process visual information in robot vision~\cite{Gallego_2020}. Unlike conventional cameras, each pixel in an event camera responds asynchronously to local brightness changes, generating a sparse stream of individual events with very low latency. Furthermore, the amount of data generated by an event camera depends on the scene dynamics. If the scene is static, no event is generated. This implies that the event camera generally generates less redundant data, therefore speeding up processing algorithms. Moreover, pixels have logarithmic response to the brightness signal, resulting in a very high dynamic range. This makes event-based sensors extremely fast, enabling them to avoid motion blur and can see dark and bright regions simultaneously. The aforementioned characteristics are appealing for a large number of systems and applications, including event-triggered control~\cite{dimarogonas2009event, jost2015optimal}, autonomous driving~\cite{lima2015clothoid}, surveillance~\cite{bozcan2021gridnet, bozcan2020air, bozcanilkercadnet}, and vision-based robot navigation~\cite{camci2019end, camci2020deep}.

\begin{figure}[t!]
\centering
\includegraphics[width=1\linewidth]{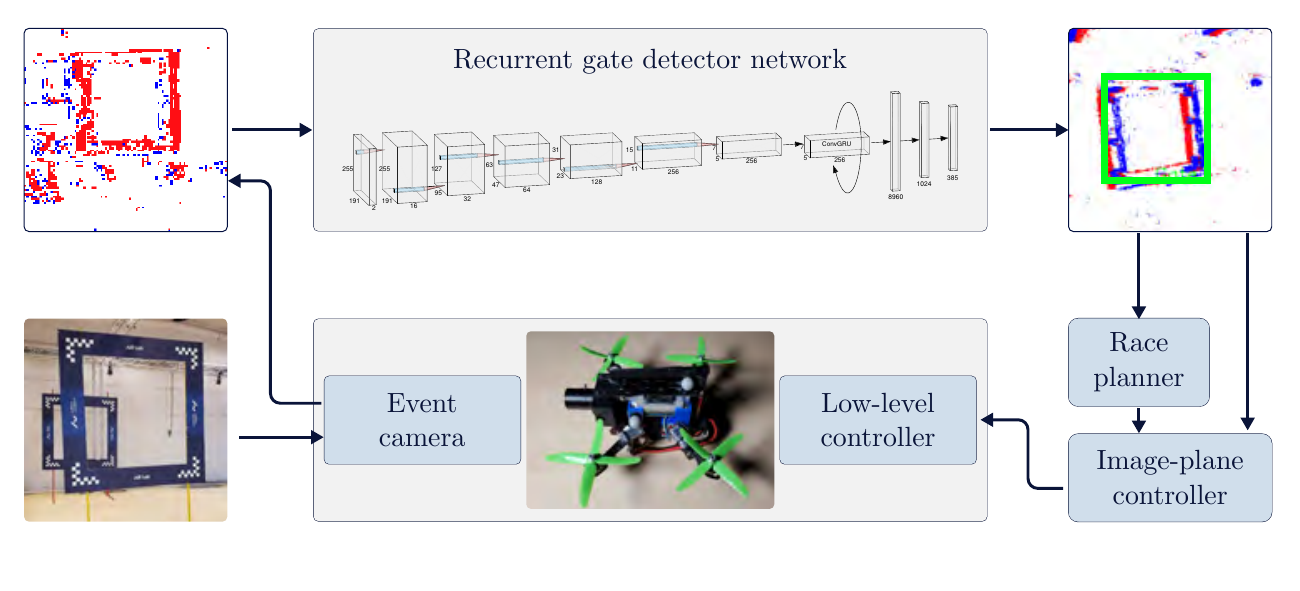}
\captionsetup{width=.9\linewidth}
\vspace{-0.75cm}
\caption{An illustration of the proposed event-based autonomous navigation and control framework. The temporal data streamed from the event camera are processed and fed into the gate detection network, which consists of a sparse backbone convolutional neural network (CNN), a gated recurrent unit, and a YOLO detection output. The predicted bounding box center of the gate is used by an image-based tracking controller to guide the drone past the gate safely. The area of the bounding box is used by a race planner to determine the next gate to pass.}
\label{fig:ev_demo}
\end{figure}

Event-based vision can be utilized to solve the challenging benchmark problem of navigation in autonomous drone racing. In this task, a flying robot is set to complete an unknown course at high speeds through a sequence of racing gates without collision~\cite{foehn2021alphapilot, pham2022deep}. A crucial aspect of this challenge is to accurately and robustly detect the drone racing gates to enable successful motion planning~\cite{huypham2021gatenet, morales2020image}. The difficulty of perceiving a particular object under aggressive and jerky maneuvers often limits the speeds and angular rates of the autonomous drone and also places a burden on controllers to ensure optimal conditions for perception~\cite{lee2020aggressive}. 

In this study, we address the gate detection task in the autonomous drone racing problem using a single onboard event camera. The motivation is to demonstrate the feasibility of using an event camera for navigation tasks and eventually heading to create agile systems with increased robustness in challenging lighting conditions and aggressive ego-motion. The contributions of this study are summarized as follows:

\begin{itemize}
    \item A novel recurrent deep learning architecture based on sparse convolutions and gated recurrent units for gate detection is proposed. We show that the recurrent unit and pretraining the network with simulation data significantly improve gate detection accuracy.
    \item A publicly available dataset using an event-based camera for the gate detection task is released, consisting of hours of both simulated and real data.
    \item The effectiveness of the event-based perception method for the autonomous drone racing problem is demonstrated in varying light conditions and high speeds through real-time experiments. 
\end{itemize}

The rest of this work is organized as follows. Section~\ref{sec:related} reviews the event-based vision methods and algorithms. Section~\ref{sec:methodology} summarizes our method for gate detection. Section~\ref{sec:experiments} compares the proposed method with a state-of-the-art baseline and presents extensive real-time experiments to demonstrate the efficiency of the proposed novel framework. Finally, some conclusions are drawn from this study in Section \ref{sec:conclusion}.

\section{Related Work}
\label{sec:related}

In order to process event-based sensor data, one method is the use of spiking neural networks (SNNs)~\cite{hats} \cite{hots} that implicitly incorporate time into neuron models. These methods preserve the temporal resolution of the data and exploit the information encoded in the timing of events. However, there are difficulties in training SNNs as well as applying them to complex tasks. Furthermore, SNNs generally achieve lower accuracy when compared to dense networks~\cite{perot2020learning}. Another approach fundamentally alters the working principle of layers of traditional convolutional neural networks (CNNs) to efficiently process asynchronous events \cite{cannici2019asynchronous} \cite{messikommer2020eventbased}, but this approach is difficult to implement. An alternative paradigm considers accumulating the event stream into a dense representation, constructing a more common input for traditional deep learning and computer vision methods. This approach tends to partially discard the temporal resolution and information but it improves spatial accuracy. Many works in this paradigm focus on developing a handcrafted or learned input representation that preserves and enhances the salient features encoded in the event data \cite{gehrig2019endtoend} \cite{AMAE} \cite{cannici2020differentiable}.\\

Since event-based object detection is a relatively new research direction in robotics, there is limited amount of studies in the literature.  Canicci et al. \cite{cannici2019asynchronous} present two network architectures inspired by the ``you only look once'' (YOLO) \cite{redmon2016you} method. One network operates synchronously on dense representations of event data, and the other operates asynchronously on raw events using a novel formalization of layers. Messikommer et al. \cite{messikommer2020eventbased} improve the results of \cite{cannici2019asynchronous} by redefining the working principle of convolutional layers to preserve the spatial and temporal sparsity better. Perot et al. \cite{perot2020learning} propose a novel recurrent architecture operating on dense input representations that learns an end-to-end memory mechanism using long short-term memory (LSTM) units. The network is demonstrated to have performance comparable to highly-tuned frame-based object detectors.

One reason for the limited number of papers in event-based object detection is the lack of datasets containing event data and ground truth annotations. Some frame-based datasets have been converted to event-based datasets, but they only correspond to short sequences with synthetic motions \cite{n_mnist_n_caltech}. Recently, large-scale datasets for autonomous driving have been published~\cite{gen1_automotive, perot2020learning} but they do not contain data collected with complex 6-DOF motions necessary for aerial robots. For this reason, in this work, we propose a novel event-based dataset intended for gate detection in autonomous drone racing.

A few papers also demonstrate the use of event cameras in aerial systems. Vemprala et al.~\cite{vemprala2021representation} present an event-based reinforcement learning network utilizing variational autoencoder for simulated quadrotors in obstacle avoidance scenarios. Salvatore et al.~\cite{salvatore2020neuro} use both SNN and conventional CNN architectures to train a reinforcement learning network with simulated event sensors obtained at high-speed for collision avoidance tasks. These results are only demonstrated through simulations, however there exists a big gap between simulations and real-world applications.

\section{Methodology}
\label{sec:methodology}
 
\begin{figure}[b]
\centering
\includegraphics[width=1\linewidth]{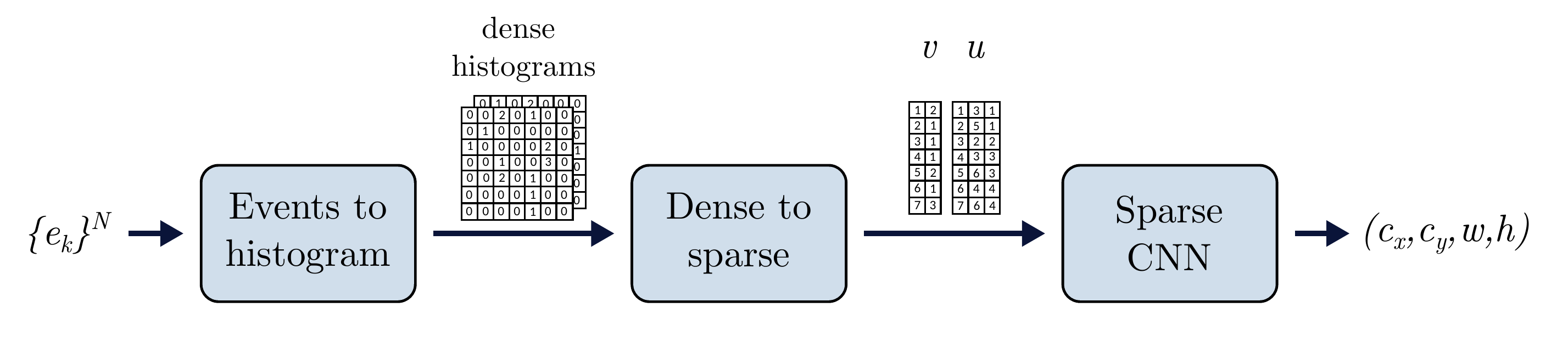}
\captionsetup{width=.9\linewidth}
\caption{Illustration of the detection pipeline. The pipeline aggregates asynchronous events into dense histograms using a fixed window and converts them into a sparse tensor of a feature vector $v$ and a location vector $u$. They are then fed into a sparse CNN network with a GRU, and regressed to obtain predictions of the center location and shape of a gate on the image plane.}
\label{fig:block_diagram}
\end{figure}

In this section, our approach towards event-based gate detection is presented. When one single pixel of an event camera detects a brightness change (i.e. due to robot motions or scene dynamics) in the scene, it immediately outputs an event $e_k$ that is a tuple containing the image plane location $x_k, y_k$, the time in microseconds $t_k$, and the one-bit polarity $p_k$ representing the positive or negative direction of the change. The overall pipeline (see Fig. \ref{fig:block_diagram}) first aggregates $N$ asynchronous events $\mathcal{E}=\left\{e_k  \right\}^{N}_{k=1}$ into two dense 2D histograms, one for each event polarity. We empirically select a fixed window $N = 10,000$ events to construct the histograms. Since event-based data are inherently spatially sparse, feeding the histograms directly into a traditional CNN is fundamentally inefficient. As the frame propagates deeper into the network, the number of active sites increases rapidly, further decreasing the network efficiency. To address this problem and preserve the spatial sparsity of event data, we utilize submanifold sparse convolutions \cite{graham20173submanifold} to redefine a wide set of CNN layers and operations optimized for sparse input data. It has been shown that these sparse convolutions achieve lower theoretical computation bound than traditional convolutions for sparse data~\cite{messikommer2020eventbased}. To make use of the new sparse CNN, the two dense histograms are converted into a sparse tensor. This tensor consists of two vectors: a feature vector $v$ containing the values of dense histograms and a location vector $u$ containing the spatial locations of each corresponding value. This process can be expressed as a mapping $\mathcal{M}$:

\begin{equation}\label{eq:histogram}
   \mathcal{M} : \mathcal{E} = \left\{e_k  \right\}^{N}_{k=1} = \left\{(x_k, y_k, t_k, p_k)\right\}^{N}_{k=1} \mapsto \mathcal{T} = (v, u).
\end{equation}
\noindent

This sparse histogram $\mathcal{T}$ in \eqref{eq:histogram} is then propagated into the backbone sparse CNN utilizing a recurrent unit. We use a YOLO~\cite{yolov1_redmon} detection output of a $5 \times 7 \times11$ tensor containing predicted bounding boxes for each grid cell in the frame. Each bounding box is of the form $B = (\hat{c_x}, \hat{c_y}, \hat{w}, \hat{h})$, representing the center location $(\hat{c_x}, \hat{c_y})$ and the size $\hat{w} \times \hat{h}$ of a gate on the image plane. The overall network architecture can be seen in Fig. \ref{fig:network_architecture}.

\subsection{Backbone CNN architecture and regression head}

In this paper, a lightweight variation of the VGG11 network \cite{vgg_simonyan2015deep} is adopted as the backbone network for feature extraction, similar to~\cite{messikommer2020eventbased}. The motivation behind this choice is that a deep neural network with small filters can learn more complex features and outperform a shallow network with large filters in many scenarios. The histogram input tensor has a spatial size of $255 \times 191 \times 2$. The network consists of 11 convolutional layers structured into blocks of two separated by a max pooling layer. All convolutions use a filter kernel size of $3 \times 3$. The first convolution layer has 16 channels, and the number of channels is doubled at each convolution block. The output of this feature extractor is a tensor of size $5 \times 7 \times 256$. 

After the last convolutional layer, the resulting tensor is flattened and fed into the regression head consisting of two fully connected layers, producing the final detection results. We apply a Rectified Linear Unit (ReLU) activation function and use batch normalization after each convolution layer to speed up training convergence. Unlike \cite{messikommer2020eventbased}, a recurrent unit is appended in-between the regression head and the backbone network in the later phase of the training.

\subsection{Gated recurrent units}

The primary reason for utilizing recurrency is to exploit the implicit relationships among event-based data at different time steps. By continuously preserving the sequential or temporal information internally as hidden states, the network can learn long-term dependencies and temporal correlations from training samples. Recurrent networks for event-based processing have earlier been utilized with success~\cite{hidalgocarrio2020learning, rebecq2019high, perot2020learning}, with \cite{perot2020learning} being the closest to this work as they tackle object detection using LSTM cells. In this paper, we choose to utilize a recent variation of a long-term recurrent cell, called the gated recurrent unit (GRU) \cite{cho2014gru}. GRUs have been shown to achieve similar performance as LSTM cells \cite{lstm_hochreiter} at a smaller computational cost since they only consist of two gates instead of three in LSTM \cite{chung2014empiricalgru}.

\begin{figure*}[hbt]
\centering
\includegraphics[width=1\linewidth]{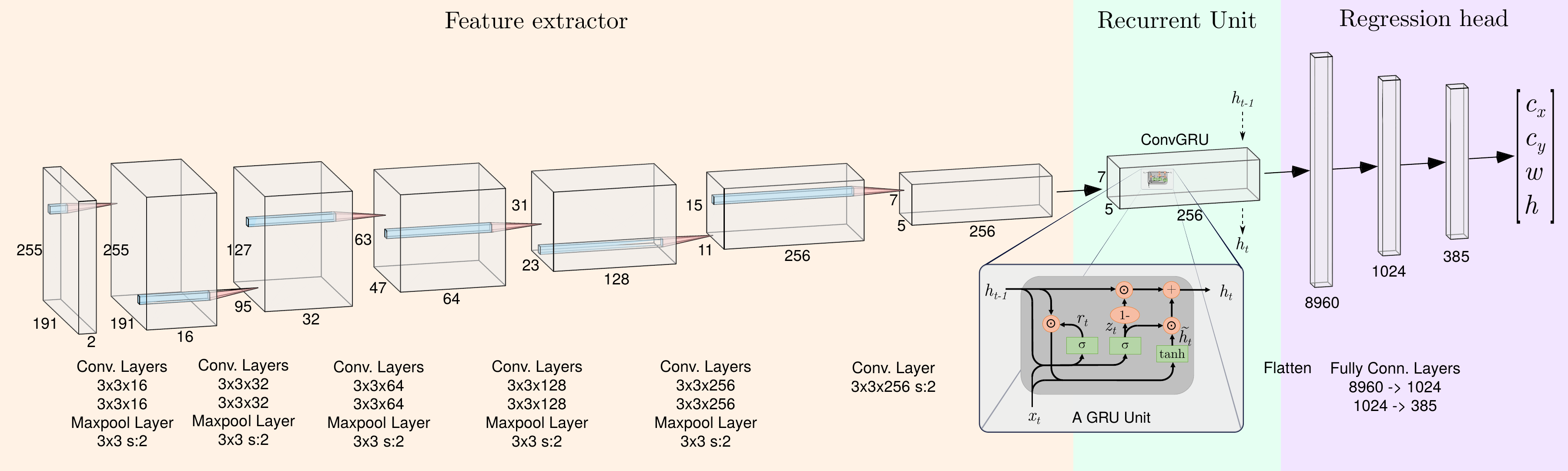}
\captionsetup{width=1\linewidth}
\caption{Proposed event-based gate detection network architecture. The network consists of (1) a backbone CNN utilizing sparse convolution as the feature extractor, (2) a GRU as the recurrent unit, and (3) a YOLO output layer as the regression head. The recurrent unit operates on both the output of last convolutional layer and the hidden state $h_{t-1}$ from the previous timestep visualized by the vertical arrows. The final vector of size 385 is reshaped into a $5 \times 7 \times 11$ tensor containing bounding boxes according to the YOLO regression method.}
\label{fig:network_architecture}
\end{figure*}

The GRU (see Fig. \ref{fig:network_architecture}) takes the feature extractor output tensor at the current time step $x_t$ and the hidden state from the previous time step $h_{t-1}$ as inputs. The update gate $z_t$ consists of a convolution with a sigmoid activation layer to determine how much information from the previous hidden state should be transferred into the new hidden state. The reset gate $r_t$ with a similar structure is used to determine how much of the past state should be forgotten. The equations governing a convolutional GRU can be expressed as:

\begin{equation}\label{eq:convgru}
\begin{aligned}
    z_t &= \sigma (W_z * [h_{t-1}, x_t]), \\
    r_t &= \sigma (W_r * [h_{t-1}, x_t]), \\
    \Tilde{h}_t &= \mathrm{tanh}(W_h * [r_t \odot  h_{t-1}, x_t]),\\
    h_t &= (1-z_t) \odot h_{t-1} + z_t \odot \Tilde{h}_t,
\end{aligned}
\end{equation}
\noindent
where $W_z, W_r, W_h$ denote trainable weights, $*$ denotes convolution, and $\odot $ denotes the Hadamard product. Equation (\ref{eq:convgru}) is implemented as a stand-alone convolutional layer, also using submanifold sparse convolutions, and incorporated into the overall CNN architecture (Fig. \ref{fig:network_architecture}), after the feature extractor, and before the regression head. It is designed to concatenate the output of the last convolutional layer with the hidden state of the previous timestep both having dimensions $7 \times 5 \times 256$, and therefore operates on a $7 \times 5 \times 512$ tensor. By placing the GRU here, we aim to model the temporal characteristics of high-level features, and at the same time, reduce the computational complexity and memory requirements of the network.

\subsection{Dataset generation}

\begin{figure}[htbp]
\centering
\includegraphics[width=0.9\linewidth]{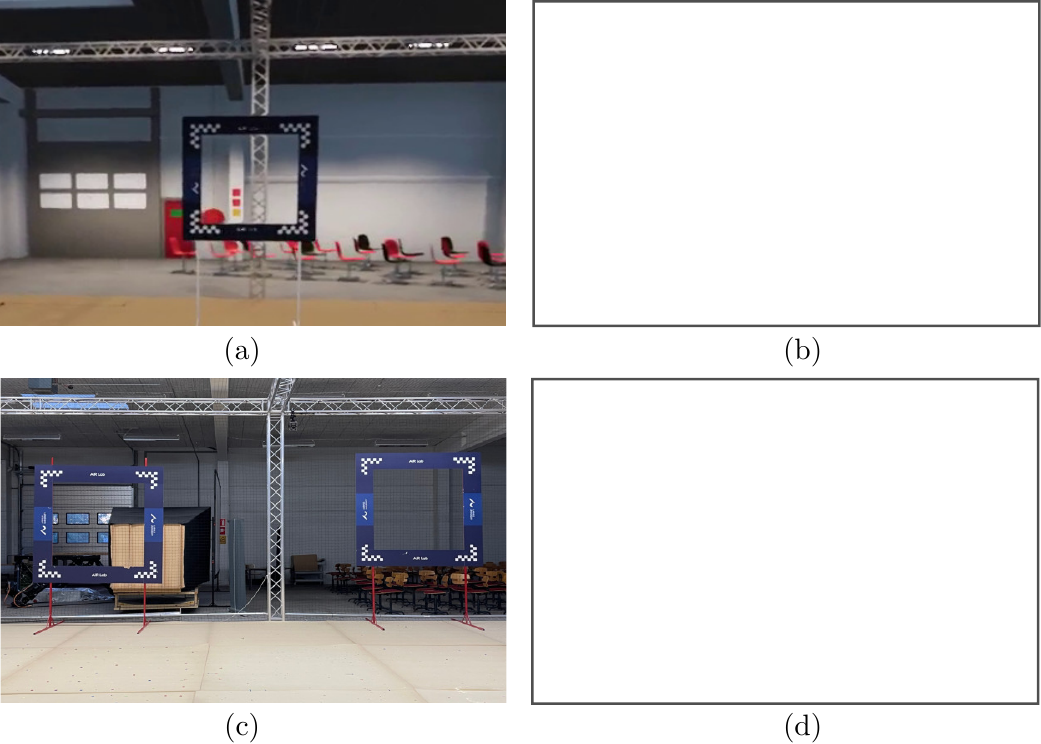}
\captionsetup{width=.7\linewidth}
\caption{Simulated and real environment. (a) An image from the simulation environment. (b) Simulated event data. (c) A photo of the real lab. (d) Real event data.}
\label{fig:sim_env}
\end{figure}

Collecting and annotating data from a real event camera is a tedious and expensive process. This work aims to reduce the use of real event data by generating event data from simulations. AirSim\footnote{\url{https://microsoft.github.io/AirSim/}}, a simulation environment based on Unreal Engine\footnote{\url{https://www.unrealengine.com/en-US/}}, is utilized to provide a physically and visually realistic simulation (Fig. \ref{fig:sim_env}). The event generation process begins with collecting RGB images from the camera of a simulated quadrotor tracking randomized trajectories at different linear and angular velocities in our simulated lab environment at the highest possible frame rate (120 \si{fps}). When a fixed number of images has been collected, the event generator ESIM \cite{ESIM} is employed to produce event data with a fixed sampling rate. Segmentation images collected from the simulation are used to automatically annotate bounding boxes around each gate. The simulation dataset contains 3.5 hours of simulated event data captured with a $240 \times 180$ sensor with over 450,000 annotated bounding boxes. 

To  evaluate the performance of the gate detection pipeline realistically, a small dataset of real event sequences is collected. We set up a quadrotor mounted with an iniVation DAVIS240\footnote{\url{https://inivation.com/wp-content/uploads/2019/08/DAVIS240.pdf}} event camera. The sensor provides a resolution of 240 x 180 pixels, with high bandwidth of 12 million events per second. 

The dataset contains sequences with varying linear and angular velocities, lab lighting, and gate placement. Afterwards, bounding boxes are manually annotated. The real dataset contains half an hour of event data captured with a real event camera and 15,000 bounding boxes. The final dataset consisting of both simulated and real sequences is named N-AU-DR and is made publicly available.

\subsection{Network training}
In order to ensure that the recurrent network learns long-term temporal dependencies, the training procedure must be carefully designed. We use truncated backpropagation through time~\cite{puskorius1994truncated} with $k = 20$ forward passes implying that the network can learn from events that happened up to 20 steps ago. The value of 20 time steps roughly corresponds to remembering 1-2 seconds back in time and was chosen empirically as it provided a decent trade-off between what memory is relevant to retain while not making training too computationally expensive. 

In practice we feed input samples to the network one time step at a time, and use a batch size of 15, meaning that we stream 15 videos simultaneously, resulting in input tensors of size $15 \times 255 \times 191 \times 2$. We do 20 forward passes while maintaining the graph and gradients, and finally do one backward pass through the entire graph to update the weights. The loss function of YOLO \cite{yolov1_redmon}, the Adam optimization algorithm \cite{kingma2014adam}, and a learning rate of $1e-4$ are employed in the training. 

In the initial experiments, the simulated data are found to be not representative enough for real-time experiments, yielding low precision scores. Instead, we use the simulated data in a transfer learning scheme aiming to help the detector generalize to different contexts. It is particularly suitable when only a small real dataset is available. The procedure for training the network is summarized in Algorithm \ref{alg:1}.

\algrenewcommand\algorithmicindent{0.3cm}%
\begin{algorithm}
\caption{\textbf{-} Recurrent neural network training procedure}\label{alg:1}

\hspace*{\algorithmicindent} \textbf{Input:} Datasets $\mathcal{X}_{sim}$, $\mathcal{X}_{real}$ and initialized model $\mathcal{M}_0$\\
\hspace*{\algorithmicindent} \textbf{Output:} Trained model $\mathcal{M}$\\
\hspace*{0.6cm}1. Train feature extractor on simulated dataset.\\
\hspace*{0.9cm}$\mathcal{M}_1 \leftarrow train\_model(\mathcal{X}_{sim}, \mathcal{M}_0)$\\
\hspace*{0.6cm}2. Train feature extractor on real dataset.\\
\hspace*{0.9cm}$\mathcal{M}_2 \leftarrow train\_model(\mathcal{X}_{real}, \mathcal{M}_1)$\\
\hspace*{0.6cm}3. Add GRU and train on real dataset.\\
\hspace*{0.9cm}$\mathcal{M}_3 \leftarrow add\_GRU(\mathcal{M}_2)$\\
\hspace*{0.9cm}$\mathcal{M} \leftarrow train\_model(\mathcal{X}_{real}, \mathcal{M}_3)$\\
\hspace*{0.6cm} \textbf{return} $\mathcal{M}$
\end{algorithm}

\section{Experiments}
\label{sec:experiments}

\subsection{Real dataset experiment}

In this part, the proposed method is evaluated for gate detection on the collected real-time dataset with various lighting and gate placements. We quantitatively compare our performance with the baseline method~\cite{messikommer2020eventbased}, as it is the only open-source state-of-the-art method that has similarities with our approach. Essentially, four different detection models are compared as follows:

\begin{enumerate}
    \item VGG Real data only: a baseline non-recurrent VGG~\cite{messikommer2020eventbased} that is trained only on the real dataset.
    \item VGG transfer learning: a baseline non-recurrent VGG~\cite{messikommer2020eventbased} that is pretrained on simulated data and fine-tuned on the real dataset. 
    \item RNN Real data only: the proposed recurrent network that is trained only on the real dataset.
    \item RNN transfer learning: the proposed recurrent network that is pretrained on simulated data and fine-tuned on the real dataset.
\end{enumerate}

The metric used is the average precision (AP), which measures precision (the number of true positives over the number of true positives and false positives) averaged across all unique recall levels.

As can be seen from Table \ref{tab:sparse_rnn_transfer_precision}, compared to the baseline method, our transfer learning method has an overall significant improvement in average precision for all real data sequences across different illumination settings. The memory mechanism of the GRU helps reduce the number of false positives (see Table \ref{tab:transfer_summary}). We do notice a significantly worse average precision when using our model on real data only, indicating that the recurrent network is more sensitive to the negative effects of only having a small train dataset. Direct sim-to-real weight transfer did not yield good results, as ESIM cannot sufficiently provide realistic event responses for the gates as its simulated output generally lacks features (e.g the checker patterns of the gates) and ambient noise, and it is generated with a much lower temporal resolution compared to a real event camera. However, simulation data helps the networks learn useful features, e.g. the outer edges of the gates.

\begin{table}[H]
\centering
\caption{Average Precision for Sparse RNN methods and the baseline methods for different sequences of the real dataset with different illumination percentages.}
\begin{tabular}{|c|ccc|}
\toprule
\backslashbox{\textbf{Model}}{\textbf{Light}} & \textbf{100\%} & \textbf{50\%} & \textbf{Varied} \\
\midrule
VGG Transfer learning~\cite{messikommer2020eventbased} & 0.455 & 0.280 & 0.117    \\ 
VGG Real data only~\cite{messikommer2020eventbased}  & 0.206 & 0.127 & 0.079   \\ 
\rowcolor{black!10} RNN Transfer learning (Ours)& \textbf{0.684} & \textbf{0.388} & \textbf{0.275}    \\ 
\rowcolor{black!10} RNN Real data only (Ours)& 0.108 & 0.040 & 0.028    \\ \bottomrule
\end{tabular}

\label{tab:sparse_rnn_transfer_precision}
\end{table}
\begin{table}[H]
\centering
\caption{Transfer learning results on average precision (AP), number of true positives (TP), false positives (FP) averaged over all real sequences.}
\begin{tabular}{|c|ccc|}
\toprule
\textbf{Model} & \textbf{AP} & \textbf{TP} & \textbf{FP} \\
\midrule
VGG Transfer learning~\cite{messikommer2020eventbased} & 0.391 & \textbf{5,569} & 5,854 \\
VGG Real data only~\cite{messikommer2020eventbased} & 0.274 & 4,044 & 5,062     \\
\rowcolor{black!10} RNN Transfer learning (Ours) & \textbf{0.567} & 5,505 & \textbf{2,442} \\ 
\rowcolor{black!10} RNN Real data only (Ours) & 0.087 & 4,787 & 35,718 \\    \bottomrule
\end{tabular}
\label{tab:transfer_summary}
\end{table}

\subsection{Real-time drone racing experiment}

\begin{figure}[t!]
\centering
\subfloat[Subfigure 1 list of figures text][]{
\includegraphics[width=0.45\linewidth]{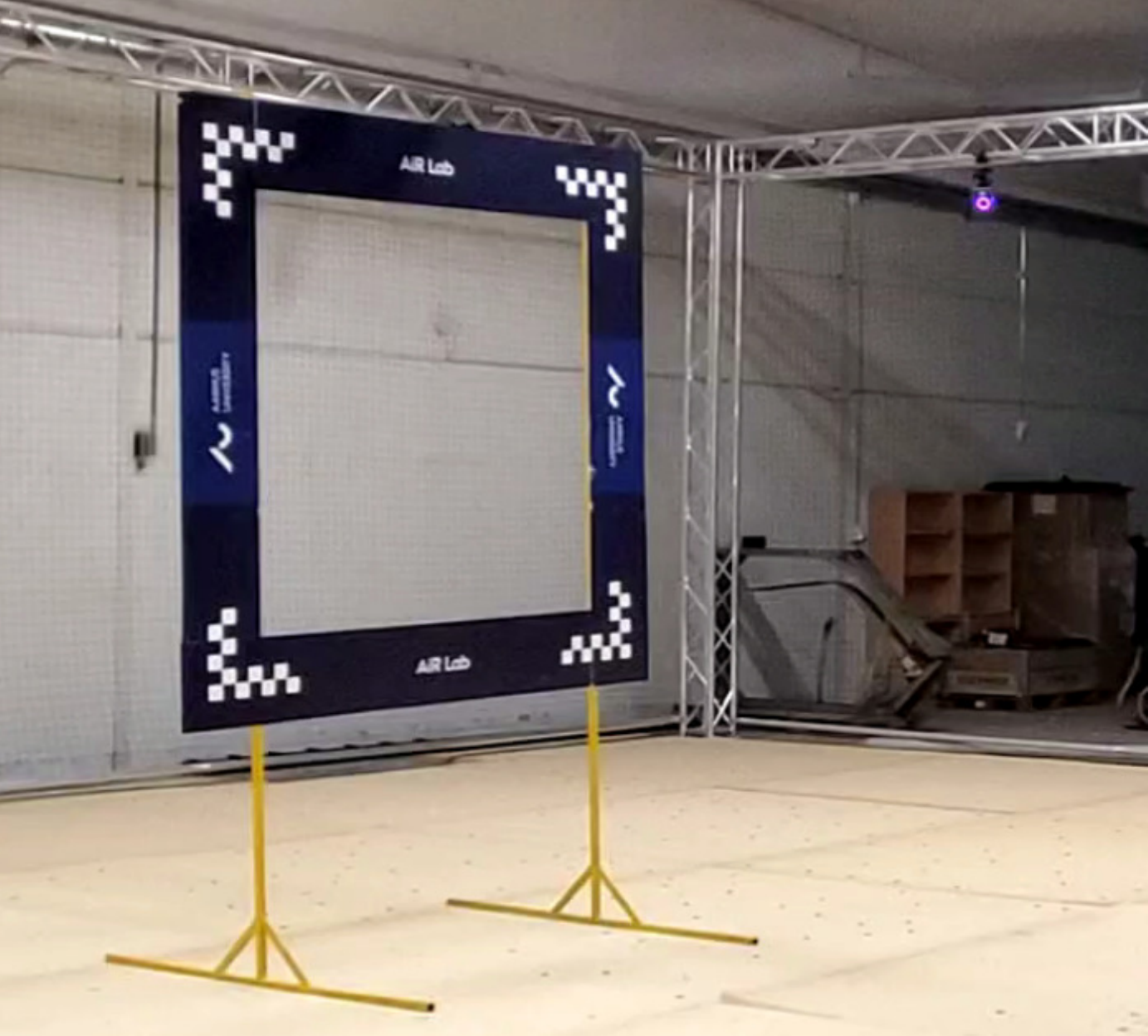}
}
\subfloat[Subfigure 2 list of figures text][]{
\includegraphics[width=0.45\linewidth]{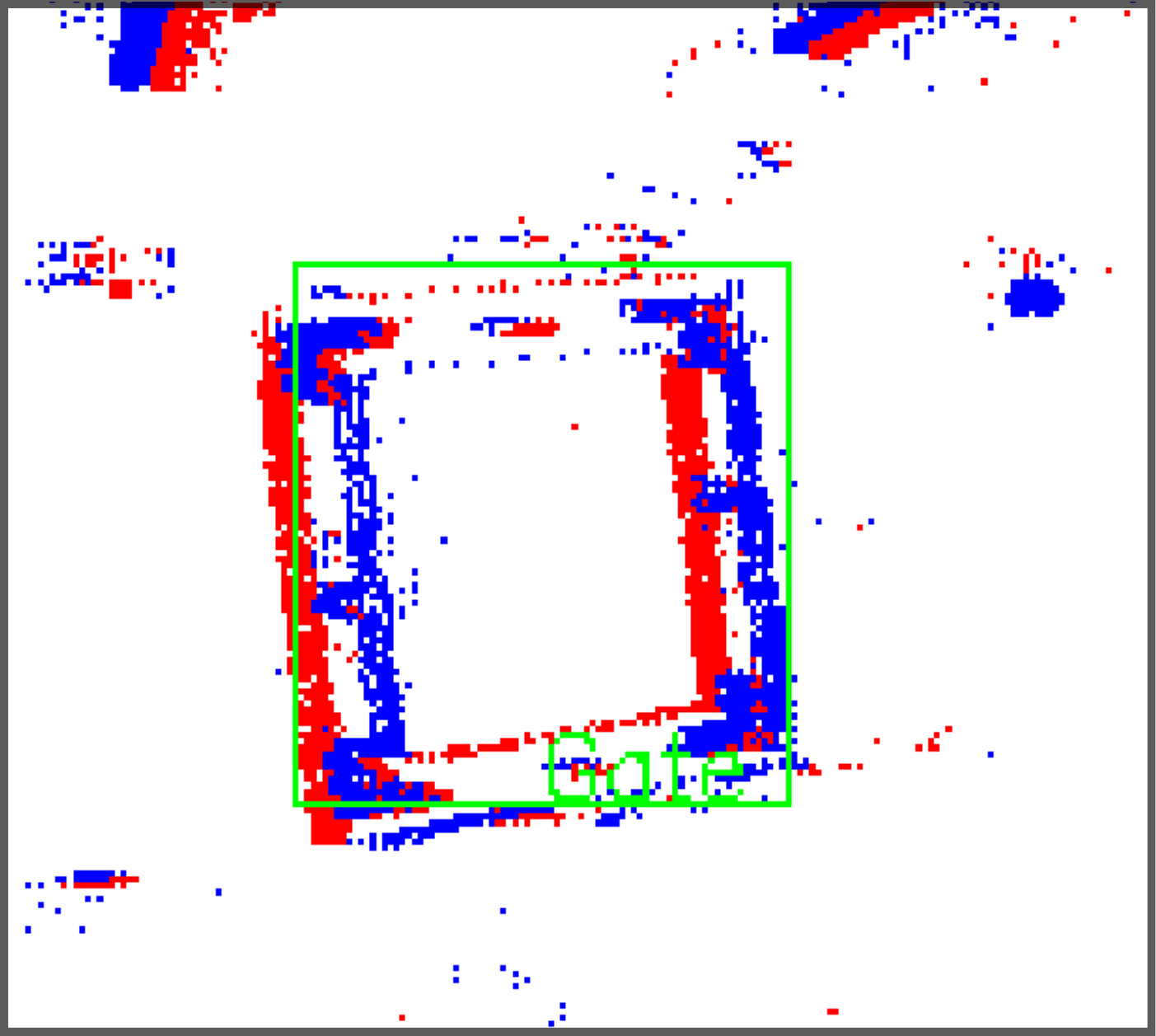}
}
\caption{(a) A typical racing gate used in the experiments. (b) A detection result and its bounding box based on event data.}
\label{fig:detction_results}
\end{figure}

Unlike some recent works in event camera-based navigation for aerial robots that show results in simulations~\cite{vemprala2021representation, salvatore2020neuro}, the viability of the proposed recurrent network method is demonstrated for a small quadrotor system to navigate in a typical real-time autonomous drone racing scenario. The quadrotor is equipped with one DAVIS240 event camera and an NVIDIA Jetson TX2\footnote{ \url{https://developer.nvidia.com/embedded/jetson-tx2}} onboard computer. It has a Pixhawk 4 Autopilot board\footnote{ \url{https://docs.px4.io/master/en/}} to handle low-level attitude and thrust control of the drone. The race track is formed by four square gates (Fig.~\ref{fig:detction_results}-(a)) with different heights but identical inner dimensions of 1.5 x 1.5 m. As the drone with propellers has a tip-to-tip diameter of 0.5 m, it is challenging for the drone to pass the gates. 

Due to the difficulties in implementing the sparse convolution network on the GPUs of the Jetson TX2, the perception pipeline inference rate is clocked at 2 \si{Hz} onboard. While this onboard system can perform some simple tasks (e.g. gate passing at slow speeds), we mostly use an offboard computer (Intel Core-i7, GPU: NVIDIA GeForce RTX 2080 Max-Q) to run the network and attain an inference rate of 40 \si{Hz} for our experiments.

In the first part of the real-time experiments, the gate detection task is demonstrated in difficult conditions. In order to recreate similar conditions in drone racing, the drone is rotated with an increasingly faster yaw rate and varying lighting conditions. A typical detection of the gate and the bounding box can be seen in Fig. \ref{fig:detction_results}-(b). Table \ref{tbl:spin_experiments} presents the successful detection rate at different speeds and scene illumination. The detection network performs well at low speeds and high illumination, except for $0.5$ rad/s and $100\%$ illumination the network sometimes misses a gate because slow movements cause the event camera to generate less data. The performance degrades at very high speeds and extreme illumination levels, which is expected due to more noisy data returned from the camera when background ambient light sources become more dominant.

\begin{table}[t!]
\caption{Success rates (in percentage) for gate detection with different angular rates and scene illumination levels.
\label{tbl:spin_experiments}}
\begin{center}
\begin{tabular}{|c|cccc|}
\toprule
 \diagbox{Spin rate}{Light} & \textbf{100}\% & \textbf{70}\% & \textbf{40}\% & \textbf{10}\% \\
\midrule
\textbf{0.5 rad/s} & \cellcolor{Green2}95  & \cellcolor{Green1}100  & \cellcolor{Green1}100  & \cellcolor{Green3}70  \\
\textbf{1.0 rad/s} & \cellcolor{Green1}100  & \cellcolor{Green1}100  & \cellcolor{Green2}80  & \cellcolor{Green4}60  \\
\textbf{2.0 rad/s} & \cellcolor{Green1}100  & \cellcolor{Green1}100  & \cellcolor{Green3}75  & \cellcolor{Green5}20  \\
\textbf{3.0 rad/s} & \cellcolor{Green2}80  & \cellcolor{Green3}70  & \cellcolor{Green4}62  & \cellcolor{Green5}20  \\
\bottomrule
\end{tabular}
\end{center}
\end{table}

In the second part, the drone is tasked to autonomously navigate through multiple unknown gates in a race track, using a single event camera onboard. The overall architecture of the system can be seen in Fig. \ref{fig:ev_demo}. Events streaming from the DAVIS camera are processed by the perception node. The outputs of this node are the predicted center of the bounding box and the size of the bounding box for each detected gate on the scene. In this work, we assume that the next gate is always visible within its field of view. The race planner node calculates the area of each bounding box and assumes the gate with the maximum area is the next gate. 

\begin{figure}[htbp!]
    \centering
    \includegraphics[width=0.49\textwidth]{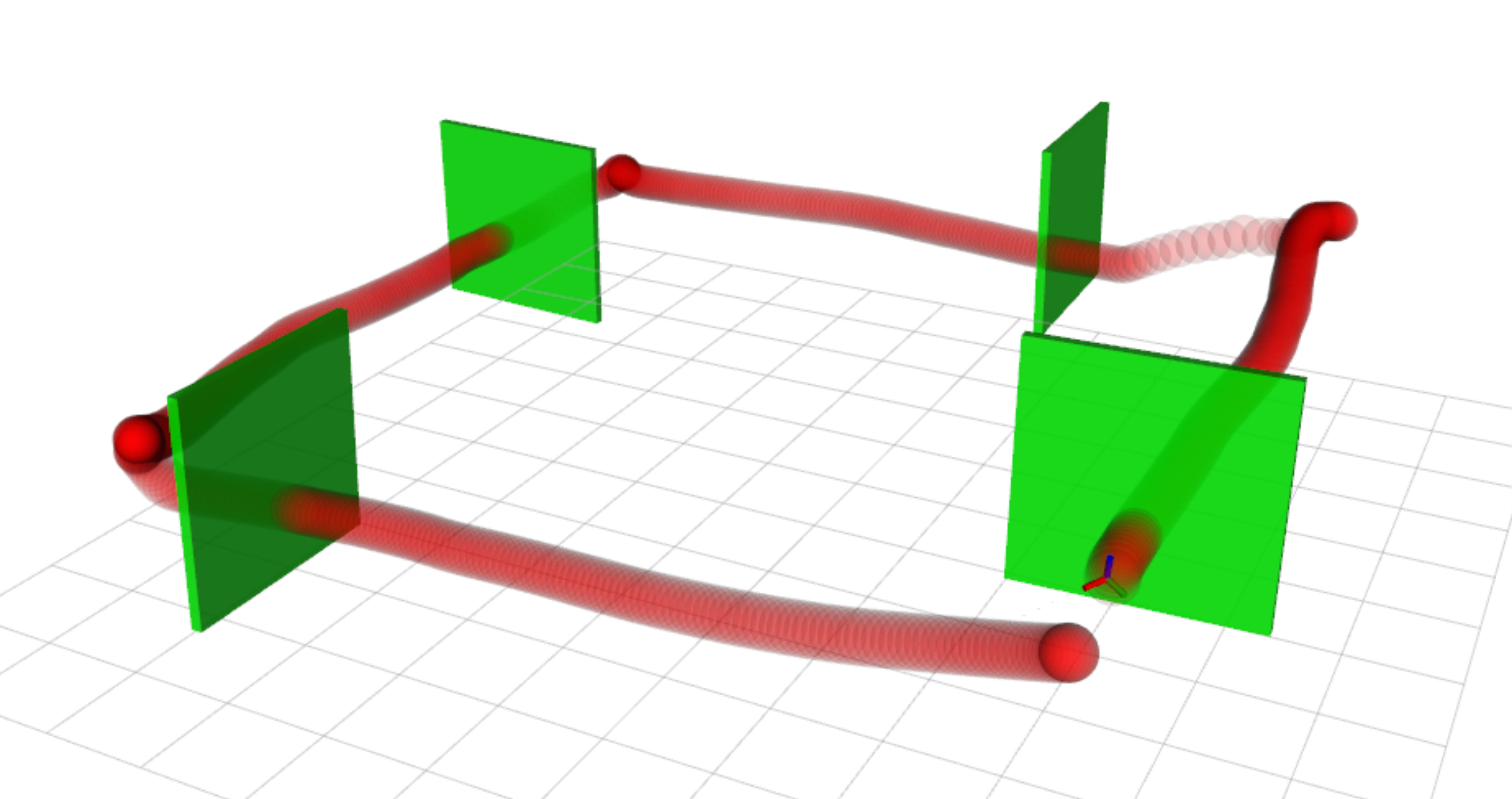}
    \caption{An example trajectory (in red color) of the drone passing multiple gates (in green color) in a race track. The video for the real-time experiments is available on the following link: \url{https://youtu.be/aFhtpOzczrc}}
    \label{fig:trajectory}
\end{figure}

To control the drone passing through a gate's center safely, we employ an image-based linear control method: as the prediction does not depend on a particular non-linear camera calibration setting, the control inputs can be considered to follow linear relationships with the lateral and vertical difference between the drone's image center $I_{c}$ and the predicted center of the gate $[\hat{c_x}, \hat{c_y}]$ on an image plane. Therefore, our controller can be expressed as:

\begin{equation}
    \centering
    \begin{cases}
    \mathbf{u} = (u_{x}, u_{y}, u_{z}),\\
    u_{x} = v_{long},\\
    u_{y} = f(I_{c,x} - \hat{c_x}) = k_{lat}\Delta_x, \\
    u_{z} = f(I_{c,y} - \hat{c_y}) = k_{ver}\Delta_y, \\
    \end{cases}
\end{equation}
\noindent
where $\mathbf{u}$ is the control input vector with component $u_x, u_y, u_z$ along the drone's body frame axes (see Fig. \ref{fig:control}), $(\Delta_x, \Delta_y)$ are differences in pixels, and $k_{lat}, k_{ver}$ are tunable gains. In the experiments, we choose a constant forward longitude velocity $v_{long} = 1$ (m/s) and only calculate lateral and vertical velocities of the drone. One advantage of this method is that the drone only requires local information from its camera and velocities on its body frame. An example of the resulting trajectories of our drone navigating a race track can be seen in Fig. \ref{fig:trajectory}. Thanks to the accuracy of the gate detection, the drone mostly tracks the gate center when passing a gate. Because of the narrow field of view of our event camera ($60^{\circ}$ horizontal x $50^{\circ}$ vertical), the drone may lose sight of the gate when it is getting close. Although this does not affect the performance in practice, as the predicted bounding boxes generally do not change in the last second before passing, the drone should have other supplement sensors to ensure safe gate passing in more complex scenarios. Readers are encouraged to see the video of the real-time experiments available on the following link: \url{https://youtu.be/aFhtpOzczrc} 

\begin{figure}[hbp!]
    \centering
    \includegraphics[width=0.49\textwidth]{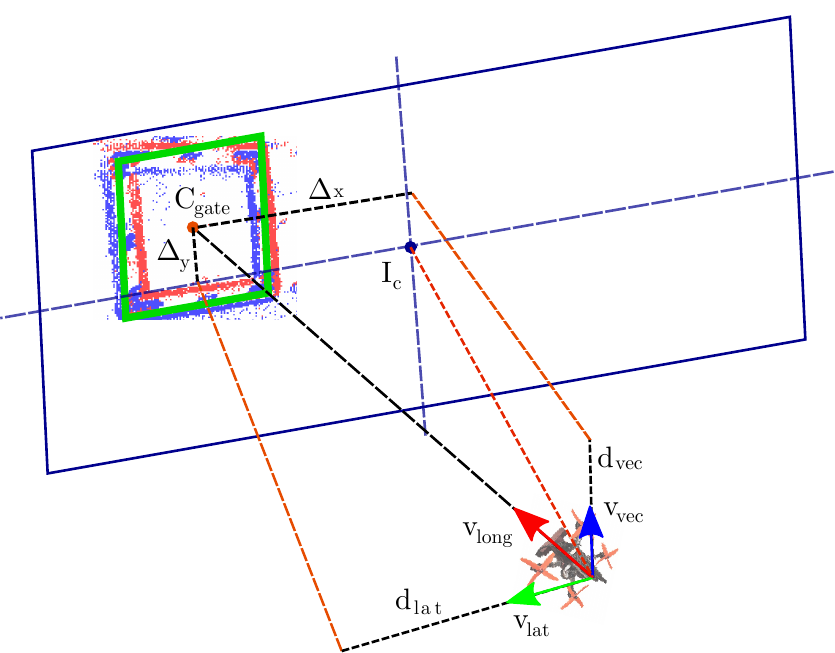}
    \caption{An illustration of the image-based linear control method.}
    \label{fig:control}
\end{figure}

\section{Conclusion}
\label{sec:conclusion}

This paper demonstrates the use of an event camera for a typical navigation and control task of an autonomous racing drone. The proposed method that relies on a sparse CNN and a GRU with sim-to-real transfer learning is shown to significantly improve the gate detection accuracy and outperforms a baseline method in our dataset experiments. Although the simulated event camera data do not realistically capture the scene information with sufficient temporal resolution, it is still helpful to pretrain the network to learn useful appearance features. The used data are organized and publicly released as a comprehensive dataset consisting of both simulated and real data for gate detection problem. The proposed method generally performs well in real-time at high speeds and under varying illumination, although it should be noted that background ambient light does negatively affect the performance at high speeds and extreme dark illumination. Finally, the  closed-loop system tests demonstrate the viability of safe navigation using a single event camera not only for autonomous drone racing, but also for other vision-based control applications. For future work, we will extend the method to predict the depth information of the gates for improved control and planning algorithms in more complicated settings, and also increase the efficiency of the network for a faster inference rate, that could allow higher drone speeds.


\section*{Acknowledgment}
The authors are grateful to Daniel Gehrig, Nico Messikommer, and Davide Scaramuzza for the fruitful scientific discussions and profound comments. This work is supported by Aarhus University, Department of Electrical and Computer Engineering (28173) and the European Union’s Horizon 2020 Research and Innovation Program (OpenDR) under Grant 871449. This publication reflects the authors’ views only. The European Commission is not responsible for any use that may be made of the information it contains.
\bibliographystyle{IEEEtran}
\bibliography{References}

\begin{thebibliography}{10}
\providecommand{\url}[1]{#1}
\csname url@samestyle\endcsname
\providecommand{\newblock}{\relax}
\providecommand{\bibinfo}[2]{#2}
\providecommand{\BIBentrySTDinterwordspacing}{\spaceskip=0pt\relax}
\providecommand{\BIBentryALTinterwordstretchfactor}{4}
\providecommand{\BIBentryALTinterwordspacing}{\spaceskip=\fontdimen2\font plus
\BIBentryALTinterwordstretchfactor\fontdimen3\font minus
  \fontdimen4\font\relax}
\providecommand{\BIBforeignlanguage}[2]{{%
\expandafter\ifx\csname l@#1\endcsname\relax
\typeout{** WARNING: IEEEtran.bst: No hyphenation pattern has been}%
\typeout{** loaded for the language `#1'. Using the pattern for}%
\typeout{** the default language instead.}%
\else
\language=\csname l@#1\endcsname
\fi
#2}}
\providecommand{\BIBdecl}{\relax}
\BIBdecl

\bibitem{Gallego_2020}
\BIBentryALTinterwordspacing
G.~Gallego, T.~Delbruck, G.~M. Orchard, C.~Bartolozzi, B.~Taba, A.~Censi,
  S.~Leutenegger, A.~Davison, J.~Conradt, K.~Daniilidis, and et~al.,
  ``Event-based vision: A survey,'' \emph{IEEE Transactions on Pattern Analysis
  and Machine Intelligence}, p. 1–1, 2020. [Online]. Available:
  \url{http://dx.doi.org/10.1109/TPAMI.2020.3008413}
\BIBentrySTDinterwordspacing

\bibitem{dimarogonas2009event}
D.~V. Dimarogonas and K.~H. Johansson, ``Event-triggered cooperative control,''
  in \emph{2009 European Control Conference (ECC)}.\hskip 1em plus 0.5em minus
  0.4em\relax IEEE, 2009, pp. 3015--3020.

\bibitem{jost2015optimal}
M.~Jost, M.~S. Darup, and M.~M{\"o}nnigmann, ``Optimal and suboptimal
  event-triggering in linear model predictive control,'' in \emph{2015 European
  Control Conference (ECC)}.\hskip 1em plus 0.5em minus 0.4em\relax IEEE, 2015,
  pp. 1153--1158.

\bibitem{lima2015clothoid}
P.~F. Lima, M.~Trincavelli, J.~M{\aa}rtensson, and B.~Wahlberg,
  ``Clothoid-based model predictive control for autonomous driving,'' in
  \emph{2015 European Control Conference (ECC)}.\hskip 1em plus 0.5em minus
  0.4em\relax IEEE, 2015, pp. 2983--2990.

\bibitem{bozcan2021gridnet}
I.~Bozcan, J.~L.~F. Sejersen, H.~X. Pham, and E.~Kayacan, ``Gridnet:
  Image-agnostic conditional anomaly detection for indoor surveillance,''
  \emph{IEEE Robotics and Automation Letters}, 2021.

\bibitem{bozcan2020air}
I.~{Bozcan} and E.~{Kayacan}, ``Au-air: A multi-modal unmanned aerial vehicle
  dataset for low altitude traffic surveillance,'' in \emph{2020 IEEE
  International Conference on Robotics and Automation (ICRA)}, 2020, pp.
  8504--8510.

\bibitem{bozcanilkercadnet}
------, ``Context-dependent anomaly detection for low altitude traffic
  surveillance,'' in \emph{2021 The IEEE International Conference on Robotics
  and Automation (ICRA)}, 2021, p. In Print.

\bibitem{camci2019end}
E.~Camci and E.~Kayacan, ``End-to-end motion planning of quadrotors using deep
  reinforcement learning,'' \emph{arXiv preprint arXiv:1909.13599}, 2019.

\bibitem{camci2020deep}
E.~{Camci}, D.~{Campolo}, and E.~{Kayacan}, ``Deep reinforcement learning for
  motion planning of quadrotors using raw depth images,'' in \emph{2020
  International Joint Conference on Neural Networks (IJCNN)}, 2020, pp. 1--7.

\bibitem{foehn2021alphapilot}
P.~Foehn, D.~Brescianini, E.~Kaufmann, T.~Cieslewski, M.~Gehrig, M.~Muglikar,
  and D.~Scaramuzza, ``Alphapilot: Autonomous drone racing,'' \emph{Autonomous
  Robots}, pp. 1--14, 2021.

\bibitem{pham2022deep}
H.~X. Pham, H.~I. Ugurlu, J.~Le~Fevre, D.~Bardakci, and E.~Kayacan, ``Deep
  learning for vision-based navigation in autonomous drone racing,'' in
  \emph{Deep Learning for Robot Perception and Cognition}.\hskip 1em plus 0.5em
  minus 0.4em\relax Elsevier, 2022, pp. 371--406.

\bibitem{huypham2021gatenet}
H.~X. Pham, I.~Bozcan, A.~Sarabakha, S.~Haddadin, and E.~Kayacan, ``Gatenet: An
  efficient deep neural network architecture for gate perception using fish-eye
  camera in autonomous drone racing,'' in \emph{2021 IEEE/RSJ International
  Conference on Intelligent Robots and Systems (IROS)}, 2021, pp. 4176--4183.

\bibitem{morales2020image}
T.~{Morales}, A.~{Sarabakha}, and E.~{Kayacan}, ``Image generation for
  efficient neural network training in autonomous drone racing,'' in \emph{2020
  International Joint Conference on Neural Networks (IJCNN)}, 2020, pp. 1--8.

\bibitem{lee2020aggressive}
K.~Lee, J.~Gibson, and E.~A. Theodorou, ``Aggressive perception-aware
  navigation using deep optical flow dynamics and pixelmpc,'' \emph{IEEE
  Robotics and Automation Letters}, vol.~5, no.~2, pp. 1207--1214, 2020.

\bibitem{hats}
A.~Sironi, M.~Brambilla, N.~Bourdis, X.~Lagorce, and R.~Benosman, ``Hats:
  Histograms of averaged time surfaces for robust event-based object
  classification,'' 2018.

\bibitem{hots}
X.~Lagorce, G.~Orchard, F.~Galluppi, B.~E. Shi, and R.~B. Benosman, ``Hots: A
  hierarchy of event-based time-surfaces for pattern recognition,'' \emph{IEEE
  Transactions on Pattern Analysis and Machine Intelligence}, vol.~39, no.~7,
  pp. 1346--1359, 2017.

\bibitem{perot2020learning}
E.~Perot, P.~de~Tournemire, D.~Nitti, J.~Masci, and A.~Sironi, ``Learning to
  detect objects with a 1 megapixel event camera,'' 2020.

\bibitem{cannici2019asynchronous}
M.~Cannici, M.~Ciccone, A.~Romanoni, and M.~Matteucci, ``Asynchronous
  convolutional networks for object detection in neuromorphic cameras,'' 2019.

\bibitem{messikommer2020eventbased}
N.~Messikommer, D.~Gehrig, A.~Loquercio, and D.~Scaramuzza, ``Event-based
  asynchronous sparse convolutional networks,'' 2020.

\bibitem{gehrig2019endtoend}
D.~Gehrig, A.~Loquercio, K.~G. Derpanis, and D.~Scaramuzza, ``End-to-end
  learning of representations for asynchronous event-based data,'' 2019.

\bibitem{AMAE}
Y.~Deng, Y.~Li, and H.~Chen, ``Amae: Adaptive motion-agnostic encoder for
  event-based object classification,'' \emph{IEEE Robotics and Automation
  Letters}, vol.~5, no.~3, pp. 4596--4603, 2020.

\bibitem{cannici2020differentiable}
M.~Cannici, M.~Ciccone, A.~Romanoni, and M.~Matteuccii, ``A differentiable
  recurrent surface for asynchronous event-based data,'' 2020.

\bibitem{redmon2016you}
J.~Redmon, S.~Divvala, R.~Girshick, and A.~Farhadi, ``You only look once:
  Unified, real-time object detection,'' in \emph{Proceedings of the IEEE
  conference on computer vision and pattern recognition}, 2016, pp. 779--788.

\bibitem{n_mnist_n_caltech}
G.~Orchard, A.~Jayawant, G.~Cohen, and N.~Thakor, ``Converting static image
  datasets to spiking neuromorphic datasets using saccades,'' 2015.

\bibitem{gen1_automotive}
P.~de~Tournemire, D.~Nitti, E.~Perot, D.~Migliore, and A.~Sironi, ``A large
  scale event-based detection dataset for automotive,'' 2020.

\bibitem{vemprala2021representation}
S.~Vemprala, S.~Mian, and A.~Kapoor, ``Representation learning for event-based
  visuomotor policies,'' \emph{arXiv preprint arXiv:2103.00806}, 2021.

\bibitem{salvatore2020neuro}
N.~Salvatore, S.~Mian, C.~Abidi, and A.~D. George, ``A neuro-inspired approach
  to intelligent collision avoidance and navigation,'' in \emph{2020 AIAA/IEEE
  39th Digital Avionics Systems Conference (DASC)}.\hskip 1em plus 0.5em minus
  0.4em\relax IEEE, 2020, pp. 1--9.

\bibitem{graham20173submanifold}
B.~Graham, M.~Engelcke, and L.~van~der Maaten, ``3d semantic segmentation with
  submanifold sparse convolutional networks,'' 2017.

\bibitem{yolov1_redmon}
J.~Redmon, S.~Divvala, R.~Girshick, and A.~Farhadi, ``You only look once:
  Unified, real-time object detection,'' 2016.

\bibitem{vgg_simonyan2015deep}
K.~Simonyan and A.~Zisserman, ``Very deep convolutional networks for
  large-scale image recognition,'' 2015.

\bibitem{hidalgocarrio2020learning}
J.~Hidalgo-Carrió, D.~Gehrig, and D.~Scaramuzza, ``Learning monocular dense
  depth from events,'' 2020.

\bibitem{rebecq2019high}
H.~Rebecq, R.~Ranftl, V.~Koltun, and D.~Scaramuzza, ``High speed and high
  dynamic range video with an event camera,'' 2019.

\bibitem{cho2014gru}
K.~Cho, B.~van Merrienboer, C.~Gulcehre, D.~Bahdanau, F.~Bougares, H.~Schwenk,
  and Y.~Bengio, ``Learning phrase representations using rnn encoder-decoder
  for statistical machine translation,'' 2014.

\bibitem{lstm_hochreiter}
\BIBentryALTinterwordspacing
S.~Hochreiter and J.~Schmidhuber, ``Long short-term memory,'' \emph{Neural
  Comput.}, vol.~9, no.~8, p. 1735–1780, Nov. 1997. [Online]. Available:
  \url{https://doi.org/10.1162/neco.1997.9.8.1735}
\BIBentrySTDinterwordspacing

\bibitem{chung2014empiricalgru}
J.~Chung, C.~Gulcehre, K.~Cho, and Y.~Bengio, ``Empirical evaluation of gated
  recurrent neural networks on sequence modeling,'' 2014.

\bibitem{ESIM}
H.~Rebecq, D.~Gehrig, and D.~Scaramuzza, ``{ESIM}: an open event camera
  simulator,'' \emph{Conf. on Robotics Learning (CoRL)}, Oct. 2018.

\bibitem{puskorius1994truncated}
G.~Puskorius and L.~Feldkamp, ``Truncated backpropagation through time and
  kalman filter training for neurocontrol,'' in \emph{Proceedings of 1994 IEEE
  International Conference on Neural Networks (ICNN'94)}, vol.~4.\hskip 1em
  plus 0.5em minus 0.4em\relax IEEE, 1994, pp. 2488--2493.

\bibitem{kingma2014adam}
D.~P. Kingma and J.~Ba, ``Adam: A method for stochastic optimization,''
  \emph{arXiv preprint arXiv:1412.6980}, 2014.

\end{thebibliography}
%

\end{document}